\documentclass[sigconf]{acmart}
\AtBeginDocument{%
  \providecommand\BibTeX{{%
    \normalfont B\kern-0.5em{\scshape i\kern-0.25em b}\kern-0.8em\TeX}}}

\usepackage{graphicx} 
\usepackage{float} 
\usepackage{subfigure} 
\usepackage{multirow}
\usepackage[normalem]{ulem}
\useunder{\uline}{\ul}{}

\usepackage{bm}
\usepackage{enumitem}
\setenumerate[1]{itemsep=0pt,partopsep=5pt,parsep=\parskip,topsep=0pt}
\setitemize[1]{itemsep=0pt,partopsep=5pt,parsep=\parskip,topsep=0pt}
\setdescription{itemsep=0pt,partopsep=5pt,parsep=\parskip,topsep=0pt}

\copyrightyear{2021}
\acmYear{2021}
\setcopyright{acmcopyright}
\acmConference[MM '21] {Proceedings of the 29th ACM Int'l Conference on Multimedia}{October 20--24, 2021}{Virtual Event, China.}
\acmBooktitle{Proceedings of the 29th ACM Int'l Conference on Multimedia (MM '21), Oct. 20--24, 2021, Virtual Event, China}
\acmPrice{15.00}
\acmISBN{978-1-4503-8651-7/21/10}
\acmDOI{10.1145/XXXXXX.XXXXXX}

\begin{document}
\fancyhead{}


\title{Robust Real-World Image Super-Resolution against\\ Adversarial Attacks}



\author{Jiutao Yue}
\authornote{Both authors contributed equally to this research.}
\affiliation{%
  \institution{Sun Yat-sen University}
  \streetaddress{}
  \city{}
  \state{}
  \country{}
  \postcode{}
}
\email{yuejt@mail2.sysu.edu.cn}

\author{Haofeng Li}
\authornotemark[1]
\affiliation{%
  \institution{Shenzhen Research Institute of Big Data, Guangdong Provincial Key Laboratory of Big Data Computing, The Chinese University of Hong Kong, Shenzhen}
  \streetaddress{}
  \city{}
  \state{}
  \country{}
  \postcode{}
}
\email{lhaof@foxmail.com}

\author{Pengxu Wei}
\authornote{Corresponding author}
\affiliation{%
  \institution{Sun Yat-sen University}
  \streetaddress{}
  \city{}
  \state{}
  \country{}
  \postcode{}
}
\email{weipx3@mail.sysu.edu.cn}

\author{Guanbin Li}
\affiliation{%
  \institution{Sun Yat-sen University}
  \streetaddress{}
  \city{}
  \state{}
  \country{}
  \postcode{}
}
\email{liguanbin@mail.sysu.edu.cn}

\author{Liang Lin}
\affiliation{%
  \institution{Sun Yat-sen University}
  \streetaddress{}
  \city{}
  \state{}
  \country{}
  \postcode{}
}
\email{linliang@ieee.org}





\begin{abstract}
Recently deep neural networks (DNNs) have achieved significant success in real-world image super-resolution (SR). However, adversarial image samples with quasi-imperceptible noises could threaten deep learning SR models. 
In this paper, we propose a robust deep learning framework for real-world SR that randomly erases potential adversarial noises in the frequency domain of input images or features. 
The rationale is that on the SR task 
clean images or features have a different pattern from the attacked ones in the frequency domain.
Observing that existing adversarial attacks usually add high-frequency noises to input images, we introduce a novel random frequency mask module that blocks out high-frequency components possibly containing the harmful perturbations in a stochastic manner. Since the frequency masking may not only destroys the adversarial perturbations but also affects the sharp details in a clean image, we further develop an adversarial sample classifier based on the frequency domain of images to determine if applying the proposed mask module. Based on the above ideas, we devise a novel real-world image SR framework that combines the proposed frequency mask modules and the proposed adversarial classifier with an existing super-resolution backbone network. 
Experiments show that our proposed method is more insensitive to adversarial attacks and presents more stable SR results than existing models and defenses.
\end{abstract}



\maketitle

\section{Introduction}
Single image super-resolution (SISR) is to recover high-resolution (HR) visual contents with clearer details and better fidelity from a degraded low-resolution (LR) image.
Image super-resolution is a fundamental problem in the field of image processing and multi-media, and has been widely investigated for a long time. Image SR algorithms could play an essential role in a variety of applications, such as multi-media processing~\cite{Jing2020light,Xiao2020space,Chen2020when,Dou2020pca}, medical imaging~\cite{lyu2020multi,chen2020joint}, depth imaging~\cite{song2020channel,ye2020pmbanet}, and remote sensing~\cite{zhang2020scene,chen2020unified}. In recent years, deep convolutional neural networks (CNNs) which consist of learnable convolution layers have demonstrated superior performance over traditional SR algorithms, due to the strong capacity of DNNs. Training CNNs usually requires a large number of paired samples which contain low-resolution input images and their corresponding high-quality version. For image SR, a straightforward way is to obtain the degraded input images by downsampling existing high-resolution images~\cite{div2k_timofte2017ntire,set5_bevilacqua2012low}. Such simulated datasets fail to model the complicated blur kernels in practical applications, which severely drops the performance of learning-based SR methods~\cite{Efrat2013Accurate,yang2014single}. To address the issue, some recent works~\cite{city100_chen2019camera,Zhang_2019_CVPR,realsr_cai2019toward,drealsr_wei2020component} have raised the real-world image super-resolution (RealSR) task and built a realistic benchmark by zooming out and in the optical lens in DSLR cameras. Deep CNNs models trained with real-world datasets are more robust to practical noises and image degeneration.

However, deep neural networks have been notoriously threatened by adversarial attacks~\cite{ifgsm_kurakin2016adversarial,Athalye2018obfuscated,univeratt_moosavi2017universal} that synthesize an adversarial sample by adding subtle noises to a natural image. Such adversarial noises are usually computed with a deliberately incorrect supervision, and the resulted adversarial sample could mislead the target neural model considerably. The same phenomenon also occurs in the single-image super-resolution task~\cite{basicatt_choi2019evaluating}. Adversarial samples could make a deep learning based SR model to predict undesirable artifacts, which implies that existing learning-based RealSR methods may lack generalization and still suffer from unknown degradation kernels. 

\begin{figure}[t]
    \centering  
    \includegraphics[width=6.5cm]{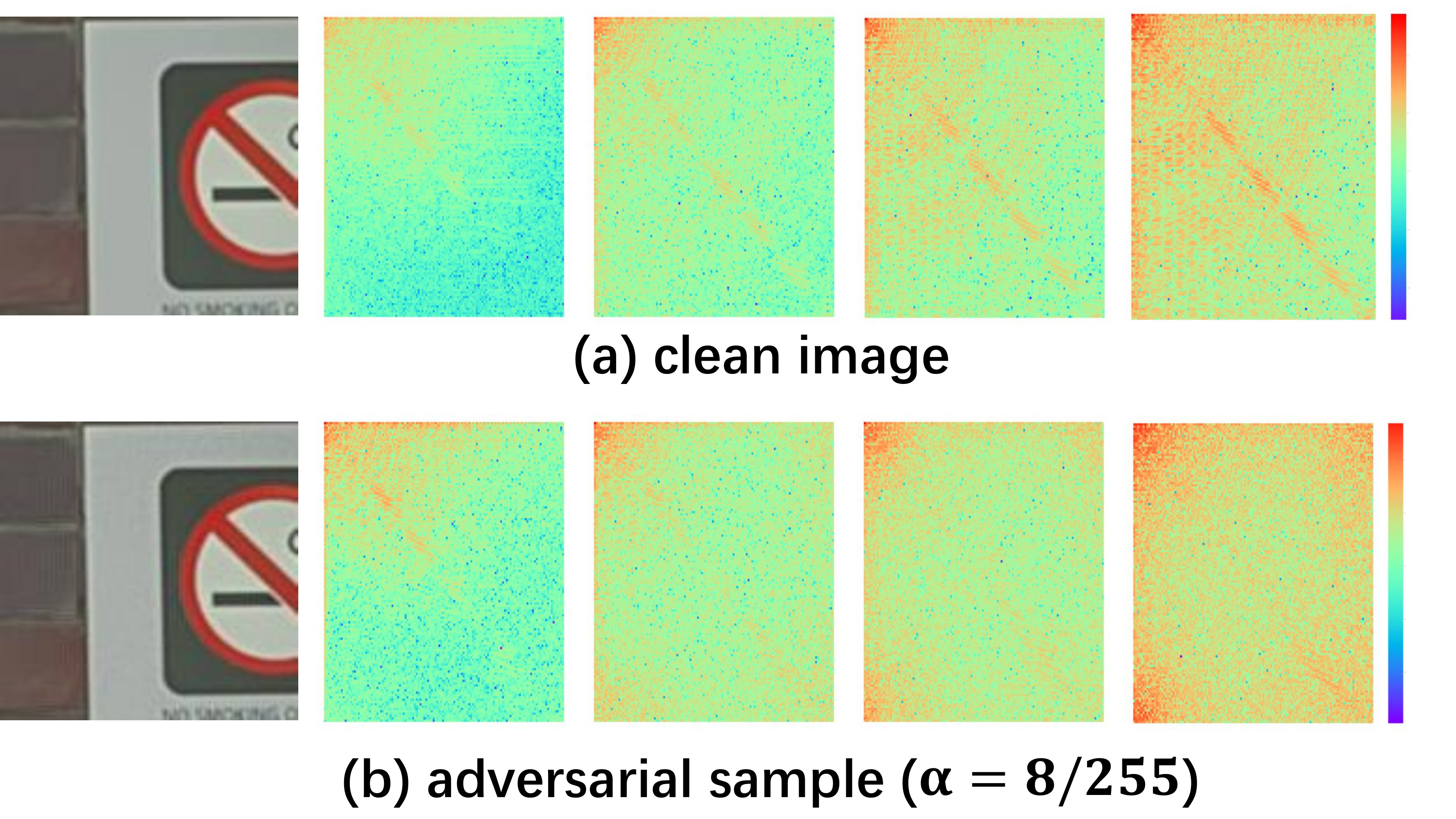}
    \caption{Visualization of 2D Discrete Cosine Transform (DCT) of the features extracted by a SR network~\cite{drealsr_wei2020component}. 
    DCT reflects the presence of signals with different frequency. In a DCT map, low-frequency signals are encoded at the upper left while high-frequency components are located at the bottom-left, bottom-right and top-right regions. In the above maps, a region in warmer color has larger values. (a) shows a clean image and the DCT maps of its neural features while (b) shows the adversarial samples and its DCTs. It can be seen that attacked features have larger values in the high-frequency components than the clean features.
    }
    \label{fig:Motivation}
\end{figure}

Most existing defenses are proposed for high-level image understanding tasks~\cite{Xie_2019_CVPR,Zhang_2019_ICCV,He_2019_nonlocal}, which may be unpromising in the low-level image super-resolution task.
In this paper, we investigate how adversarial perturbations affect SR models from a frequency perspective, by visualizing the frequency domain (\textit{e.g.} Discrete Cosine Transform, DCT) of image features as shown in Figure~\ref{fig:Motivation}. We surprisingly find that adversarial attacks did change the DCT pattern of an image feature map extracted by deep SR models. For a natural image, its frequency domain map usually contains the most significant values in low-frequency coefficients which encode the flat image regions of similar colors. The middle-to-high frequency components typically have the second largest values, which reflects the presence of sharp edges and corners. The smallest coefficients usually occur at the high-frequency component which is related to the highly repetitive elements in an image, such as noises, texture and artifacts. Interestingly, attacking a SISR model considerably increases the high-frequency coefficients from either vertical or horizontal direction in the attacked image features, which corresponds to densely distributed adversarial noises. The DCT difference between clean features and the attacked ones provides a hint to detect and resist adversarial samples.

Motivated by the above observations, we propose to improve the robustness of SR neural networks with a frequency mask module that reduces adversarial noises in the level of frequency domain.
Considering that adversarial noises are parts of the high-frequency elements in an image, as shown in Figure~\ref{fig:Motivation}. The proposed mask module fills zeros in the high-frequency components for the DCT of images or features to destroy the adversarial perturbations. 
On the other hand, an image may contain highly-repetitive textures, which are encoded as high-frequency parts as the noises. It is hard to separate the harmful noises from natural textures in a frequency domain. We find that a component of higher frequency is more likely to encode adversarial noises, and design a probability distribution to randomly determine if setting a coefficient as zero. Such an improvement could better preserve the original contents of the input image. 
For a clean image, it is undesirable to discard its high-frequency elements which may include finer details of the image. Thus we further devise an image classifier based on the observation that adversarial samples have a different distribution of the frequency domain from the original images. The proposed classifier takes a visualized frequency domain map as input and predicts if the corresponding image is an adversarial sample. If not, our proposed frequency mask module could be skipped so that the sharpness and fidelity of clean images are well maintained. Then we deploy the random frequency mask module and the adversarial sample classifier to an existing neural network to construct a robust super-resolution model. 
In summary, our main contributions have three folds:
\begin{itemize}
\item We propose a random frequency mask module that erases the high-frequency components of input images and features with a prior probability distribution to mitigate adversarial attacks.
\item We introduce a frequency-based adversarial sample classifier which determines if applying the proposed mask module and helps maintain sharp details for clean images.
\item We develop a robust image super-resolution network with the proposed frequency mask module and the adversarial sample classifier. The proposed method not only achieves satisfactory super-resolved results on real-world images, but also obtains the state-of-the-art performance against adversarial attacks.
\end{itemize}

\vspace{-1em}
\section{Related Work}
\subsection{Real-World Image Super-Resolution}
Single-image super-resolution (SISR) requires to synthesize high-resolution contents from a single low-resolution image. SISR algorithms have been studied for a long time, and can be grouped into two categories: traditional and learning-based methods. Traditional SR methods are mainly based on edge priors~\cite{Raanan2007image,sun2008image}, image registration~\cite{irani1991improving} and statistics~\cite{huang1999statistics,shan2008fast,Aly2005image}. Among learning-based SR methods~\cite{Kim2010single,he2011single,ni2007image,yang2010imagesr}, DNN~\cite{dong2016imagesr,zhang2017beyond,lai2017deeplap} is one of the most popular and effective models.
According to the source of degraded images, SISR tasks could be divided into three groups: non-blind synthetic SR, blind image SR and real-world SR. Non-blind synthetic SR~\cite{dai2019second,guo2020closed} simply adopts bicubic or Gaussian downsampling to simulate the image degradation on both training and testing set, but the SR models trained with the synthetic dataset may fail to adapt to unknown blur kernels in practice.
Blind image SR~\cite{gu2019blindsr,huang2021fast} is based on a more realistic setting that the blur kernels during the testing are unavailable when training a SR model. Blind image SR algorithms could be evaluated with synthetic or real image sets, but these real datasets lack HR ground truths.

To fill the gap, real-world image SR task~\cite{city100_chen2019camera,Zhang_2019_CVPR,realsr_cai2019toward,drealsr_wei2020component} has been proposed, upsampling real degraded images by collecting many pairs of real LR sample and its HR version.
Cai \textit{et al.}~\cite{realsr_cai2019toward} employ two types of digital cameras to collect real SR pairs as a new dataset, RealSR. 
Wei \textit{et. al.}~\cite{drealsr_wei2020component} further provide a larger real-world SR benchmark DRealSR, which consists of well-aligned SR pairs captured by up to five different DSLR cameras. 
Particularly, Cai \textit{et al.}~\cite{realsr_cai2019toward} reveal that existing SR networks trained on a simulated dataset~\cite{div2k_timofte2017ntire} do not show advantages with real-world samples. It is because real degradation includes lots of factors, \textit{e.g.}, anisotropic blur, signal-dependent noise and cross-camera degradation processes. Thus real-world image SR is a more challenging and meaningful problem.

To solve this task, Cai \textit{et. al.}~\cite{realsr_cai2019toward} develop a laplacian pyramid kernel prediction network LP-KPN to explicitly learn a specific restoration kernel for each pixel.
Wei \textit{et. al.}~\cite{drealsr_wei2020component} introduce a component divide-and-conquer model CDC that constructs three Component-Attentive Blocks (CAB) associated with flatten regions, edges, and corners. CDC infers super-resolved contents with the outputs of three CABs, and obtains the state-of-the-art results in real-world image super-resolution. However, these real-world SR networks may still be sensitive to adversarial noises, even though they are trained with complicated blur kernels. In this paper we focus on implementing a robust neural network for real-world SISR.

\subsection{Adversarial Attacks and Defenses}
Previous studies have shown that DNNs are vulnerable to adversarial attacks~\cite{optimatt_szegedy2013intriguing,univeratt_moosavi2017universal,Rahmati2020geoda} which cheat a neural model by applying inconspicuous changes to an input image. 
Adversarial attacks can be categorized into two groups: black-box and white-box attacks, according to the knowledge acquired by attackers. Black-box attackers~\cite{ilyas2018black,Papernot2017practical} are allowed to access only limited knowledge of the data and the targeted network, such as the output of querying the targeted model. In this paper we only consider white-box attacks~\cite{optimatt_szegedy2013intriguing,He_2019_parametric} in which all parameters of the target model are exposed to the attacker.
As for white-box attacks, Szegedy \textit{et al.}~\cite{optimatt_szegedy2013intriguing} for the first time propose an adversarial attack for deep learning models, by maximizing the classification loss and minimizing the magnitude of the adversarial perturbation with a box-constrained L-BFGS.
Goodfellow \textit{et al.}~\cite{fgsm_goodfellow2014explaining} introduce an attack, fast gradient sign method (FGSM), which computes backward propagated gradients to maximize the classification loss and takes the sign of gradients to update an adversarial sample. Kurakin \textit{et al.}\cite{ifgsm_kurakin2016adversarial} further extend FGSM to an iterative variant, I-FGSM. To attack an image SR network, Choi \textit{et al.}~\cite{basicatt_choi2019evaluating} respectively implement basic, universal and partial attacks based on I-FGSM.
The basic and the universal attacks can affect existing CNN-based image super-resolution models considerably. Following Choi \textit{et al.}~\cite{basicatt_choi2019evaluating}, we adopt the basic attack in this paper since it presents higher success rate of attack than the universal one.

Many defense methods and robust models~\cite{Prakash2018deflecting,Choi_2020_ACCV,li2020online,zhou2021ssmd} which have been developed to resist adversarial attacks, attempt to remove, destroy or adapt to adversarial noises. Adversarial training~\cite{tramer2018ensemble,freeat_shafahi2019adversarial,zheng2020efficient} is a large group of defense methods that trains the target network with adversarial samples and are effective against white-box attacks.
Some early defenses that apply image transformations~\cite{guo2018countering,xie2017mitigating,li2020rosa} to disrupt or eliminate adversarial noises, are limited to gray-box setting~\cite{Athalye2018obfuscated} where the defense is unknown for attackers. Other improved methods that propose new neural modules to denoise or smooth image features~\cite{Xie_2019_CVPR,He_2019_nonlocal,Li_2020_CVPRenhancing}, do show their robustness. Zhang \textit{et al.}~\cite{zhang2019adversarial} propose to suppress high frequency in the discrete Fourier transform (DFT) of an input image, which is mostly related to our proposed method. Zhang \textit{et al.}~\cite{zhang2019adversarial} utilize a fixed radius to erase noises, which discarded all the high-frequency elements including natural textures in an image. Differently, we attempt to maintain some parts of high-frequency components by randomly masking the perturbations, so that our proposed method has the chance to reconstruct the original high-frequency textures.

\section{Methodology}
In this section, we first propose a novel random frequency mask module on the basis of discrete cosine transform (DCT) and Bernoulli distribution. Then we introduce an image classifier to detect adversarial samples according to the DCT pattern of the input image. Subsequently, we discuss how to incorporate the mask module and the classifier into an existing super-resolution backbone to build our proposed robust SR network. At the end of this section, we brief how to tune the overall network with the frequency mask modules and the adversarial classifier via an adversarial training strategy.

\subsection{Random Frequency Mask}
\begin{figure}[!t]
	\centering
	\includegraphics[width=7.5cm]{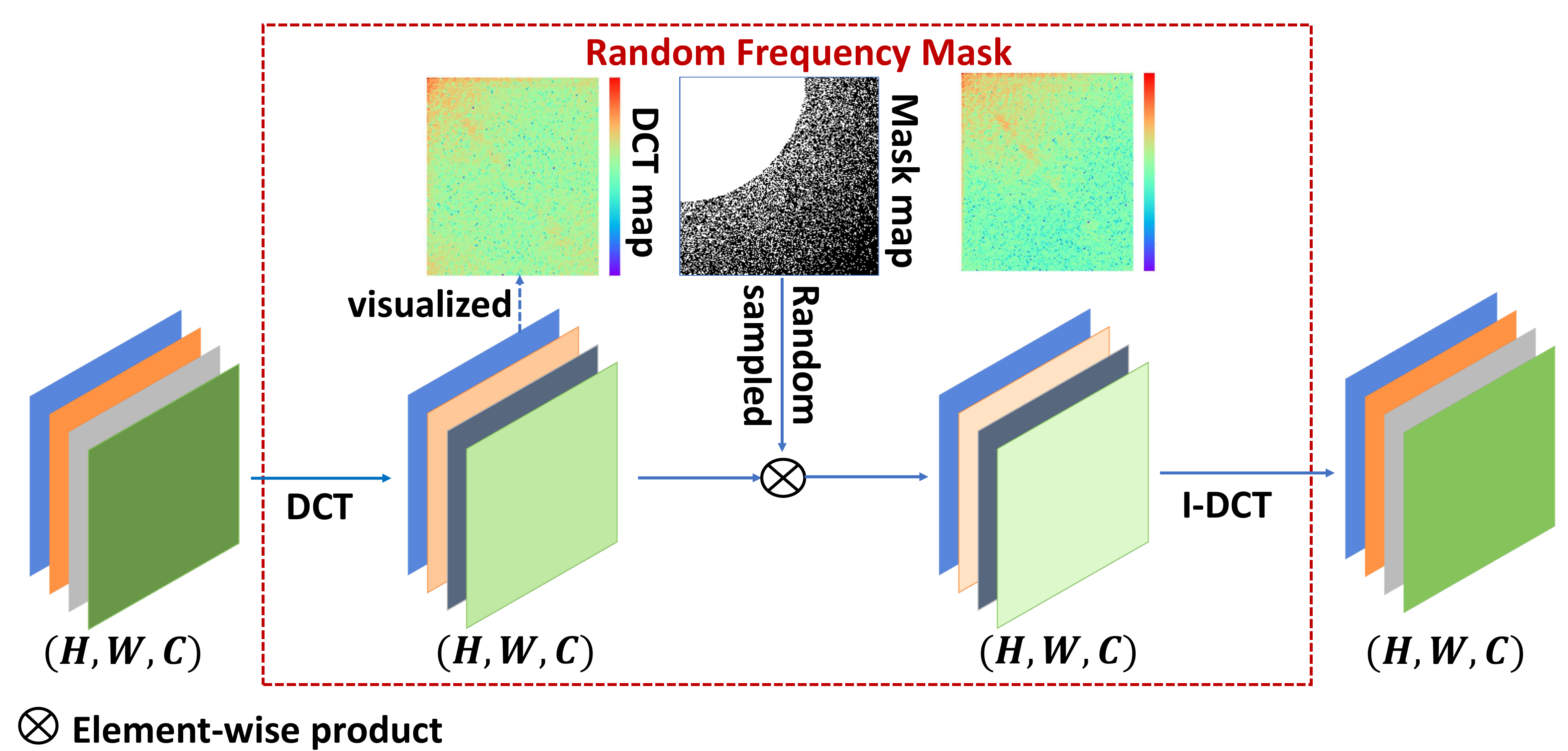}
	\caption{The architecture of our proposed random frequency mask module. The input of the mask module could be an image or a feature map extracted by super-resolution networks. First, the input image or feature is mapped into a frequency domain by applying discrete cosine transformation to each channel slice. Then a sector mask is synthesized based on a Bernoulli distribution, and is used to element-wisely multiply with the DCT maps to lower potential adversarial noises. Afterwards, the output feature of our proposed module is obtained by performing an inverse-DCT operation on the masked DCT maps.}
	\label{fig:Mask}
	\vspace{-1em}
\end{figure}
We develop a novel random frequency mask module to mitigate adversarial attacks, since most of adversarial perturbations are encoded as high-frequency components in a frequency domain. Figure~\ref{fig:Mask} illustrates the architecture of the proposed mask module that consists of three steps: transforming an input to the DCT frequency domain, masking the DCT representations with a sampled map, and converting the masked DCT back to the image or feature space.

Our proposed mask module could take an image or a feature map as input. Without loss of generality, the module input is denoted as a $H\times W\times C$ tensor.
Let $X \in \mathcal{R}^{H \times W}$ denote one of the $C$ channel slices in the $H\times W\times C$ input tensor. 
We adopt discrete cosine transform to compute the frequency representations of $X$. Let $\hat{X} \in \mathcal{R}^{H \times W}$ stand for the DCT result of $X$. $\hat{X}$ can be calculated as:
\begin{equation}
\begin{split}
    \hat{X}(u, v)=c(u) c(v) \sum_{i=0}^{H-1} \sum_{j=0}^{W-1} \Bigg\{ X(i, j) &\cos [(i+0.5)\pi /H \cdot u] \\
    &\cos [(j+0.5)\pi /W \cdot v]
\end{split}
\label{Mask_dct1}
\end{equation}
where $c(u)$ is a compensation coefficient. $c(u)$ is set as $\sqrt{1/H}$ for $u=0$ and $\sqrt{2/H}$ for $u\neq 0$, and the definition of $c(v)$ is the same as $c(u)$.

To understand DCT representations, we normalize and take average of different channels in the DCT results that are computed from a clean or attacked input. The averaged DCT maps are visualized in Figure~\ref{fig:Motivation} where the first row is for clean inputs and the second is for the attacked ones. In a DCT map, low-frequency coefficients are located nearby the upper-left corner. Figure~\ref{fig:Motivation} shows that the attacked DCT maps have larger high-frequency components, and smaller middle-frequency coefficients than the clean ones. We argue that adversarial attacks harm the original middle-frequency details to some degree, and add more high-frequency noises to the attacked image. Thus we propose to reduce the adversarial perturbations by multiplying the DCT ${\hat{X}}$ with a binary mask $\mathcal{M} \in \mathcal{R}^{H \times W}$: $\hat{X}_{m} =  {\hat{X}} \odot {\mathcal{M}}$,
where $ \odot$ is element-wise multiplication. Then the masked DCT $\hat{X}_{m}$ is translated to the same image or feature space as the module input $X$ via inverse discrete cosine transform. The whole process of our proposed mask module is formulated as: $X_m = \mathcal{F}^{-1}(\mathcal{M} \odot \mathcal{F}(X))$, 
where $\mathcal{F}(\cdot)$ stands for DCT, and $\mathcal{F}^{-1}$ denotes the Inverse-DCT. The module output $X_m$ has the same shape as the input $X$.

We discuss how to determine the binary mask $\mathcal{M}$. Considering that the distance from the lowest-frequency component corresponds to the frequency degree of a component. For each coefficient of position $(u,v)$, We set its corresponding weight in the binary mask according to its normalized distance from $(0,0)$: 
\begin{equation}
r_{(u,v)}= \sqrt{u^{2}+v^{2}} \big/{r_{max}}
\label{Mask_ruv}
\end{equation}
where $r_{max}$ equals to $\sqrt{(H-1)^{2}+(W-1)^{2}}$ and denotes the maximum radius for a DCT map of size $H\times W$. 
For a DCT component whose $r_{(u,v)}$ is smaller than a threshold $r_t$, the component probably contains the information of mean colors, smooth regions or sharp edges, which are image elements of low-to-middle frequency. To preserve the original image contents, we keep such a DCT coefficient unchanged by setting $\mathcal{M}(u,v)$ as 1. Since the boundary between middle-frequency details and high-frequency noises is uncertain, we uniformly sample $r_t$ from $[r_l, r_u]$. $r_l$ and $r_u$ are manually set lower and upper bounds respectively by visualizing the difference of DCT maps between clean and attacked samples.
If the $r_{(u,v)}$ value of a DCT coefficient is larger than the threshold, the coefficient might still encode normal contents such as highly repetitive textures. Since a higher-frequency coefficient more possibly contains adversarial noises, we adopt a Bernoulli distribution with the probability $p=r_{(u,v)}$ to decide if masking the current component. The distribution returns 1 with probability p and 0 with probability 1 - p. The binary mask $ \mathcal{M}(u,v) = Bernoulli(p=r_{(u,v)})$ is formally defined as:
\begin{equation}
    \mathcal{M}_{(u,v)}=\left\{\begin{array}{lc}
    1, \qquad\qquad\qquad\qquad\quad 0\leq &r_{(u,v)} \leq r_t \\
    Bernoulli(p=r_{(u,v)}), & r_{(u,v)} > r_t
    \end{array}\right.
\label{Mask_map}
\end{equation}
The proposed mask module has two strengths. First, it supports backward propagation and could be placed at arbitrary positions of a SR network. Second, the mask module has no learnable parameters and is a lightweight module.

\begin{figure}[!tb]
	\centering  
	\includegraphics[width=7.5cm]{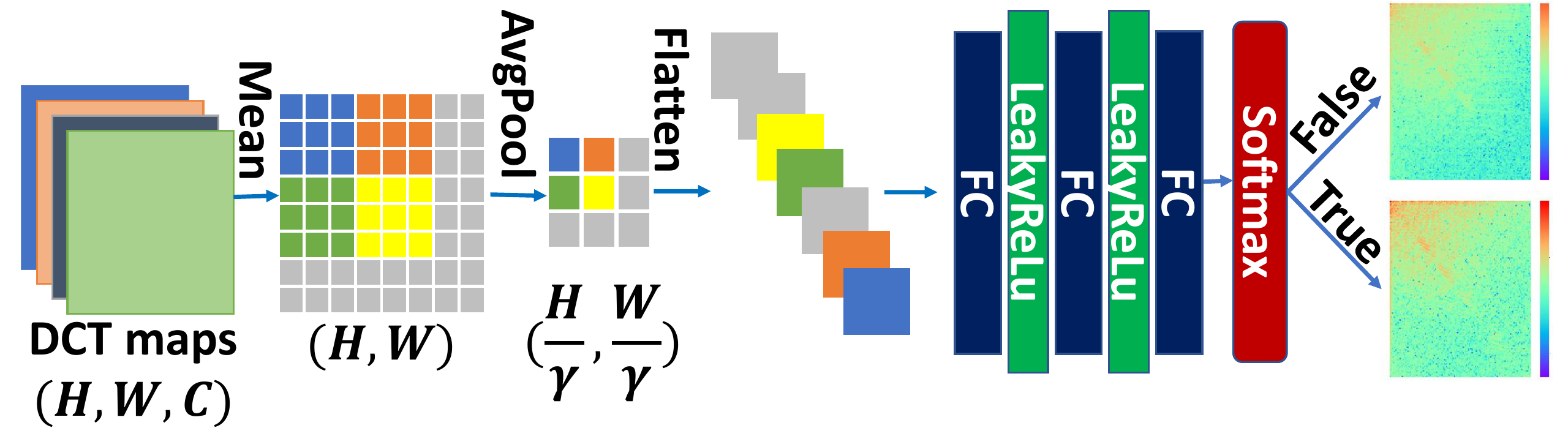}
	\caption{The architecture of our proposed adversarial sample classifier. The classifier takes DCT maps of an image as input, pools and reshapes them into a vector that is fed to a two-class classifier with three fully-connected layers. $\bm{True}$ denotes the input sample is attacked while $\bm{False}$ means a clean input. $\gamma$ is a scaling factor.}
	\label{fig:Classifier}
\end{figure}

\subsection{Adversarial Sample Classifier}
\begin{figure*}[!htb]
	\centering
	\includegraphics[width=15cm]{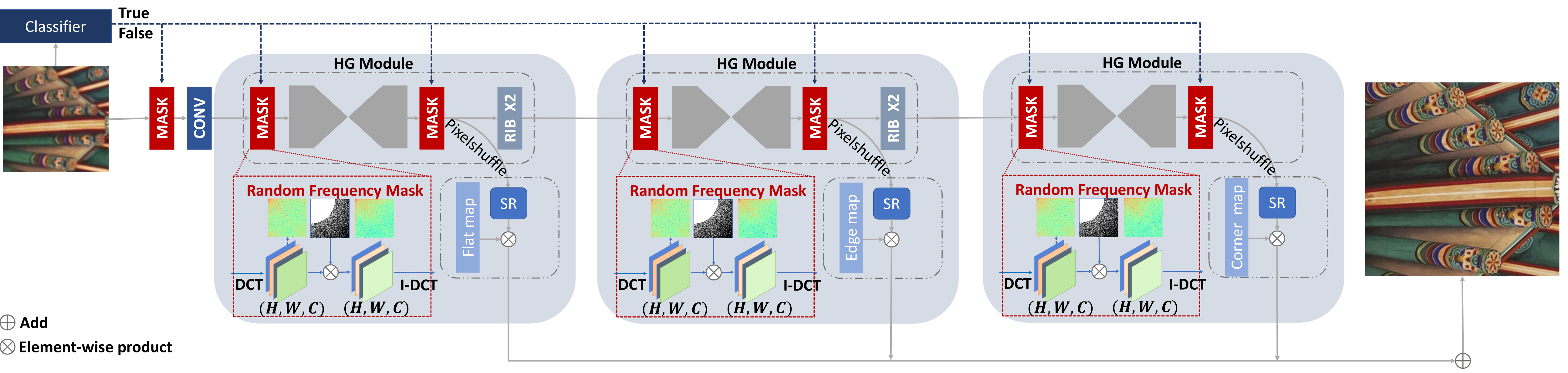}
	\caption{The overview framework of our proposed SR network. We equip a existing baseline CDC~\cite{drealsr_wei2020component} with our proposed defenses. The proposed classifier and a mask module are located at the head of the network. Two mask modules are placed at the head and the tail of each hourglass (HG) module. Each HG module has a encoder-decoder architecture with skipped residual connections. Only if the classifier predicts the input as an adversarial sample, the proposed mask modules are activated.}
	\label{fig:Network}
\end{figure*}
Although we have developed a stochastic strategy to eliminate adversarial noises and preserve original contents at the same time, the proposed mask module still inevitably discards some fine details and degrades the performance on clean images. Thus we devise a two-class classifier shown in Figure~\ref{fig:Classifier} to predict if the input image is adversarial. The frequency mask modules are skipped for a clean input so that the high-frequency elements of clean images are not affected. Considering the pattern difference between clean DCT maps and the attacked ones, we take DCT results as the classifier input.

\begin{figure*}[!tb]
	\centering
	\includegraphics[width=16cm]{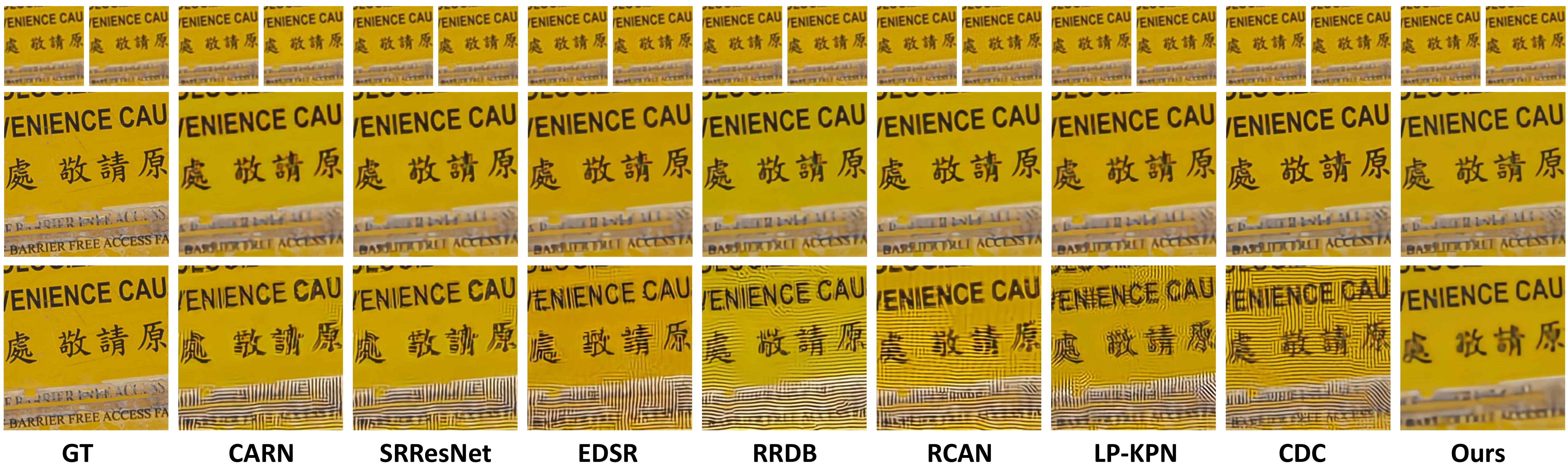}
	\caption{Super-resolved results of our proposed method and existing SR methods on the RealSR dataset.
		The leftmost column shows the ground truth. For other columns, the left of the first row is a clean image, while the right image is the adversarial sample with the intensity $\alpha = 8/255$. The middle row is the super-resolved output of the clean images. The bottom row corresponds to the super-resolved results of the adversarial samples.}
	\label{fig:Sota_image}
\end{figure*}

For an input image $X \in \mathcal{R}^{H \times W \times C}$ that could be an adversarial sample or a clean one, we first compute its DCT of the same shape of $H\times W\times C$. Then we take average of different channel slices to obtain a single-channel DCT map of size $H\times W$, which corresponds to the visualization in Figure~\ref{fig:Motivation}. The DCT map is further processed by a $\gamma\times\gamma$ 2D average pooling layer with stride 3 and padding 0. The shape of the pooled map becomes $H/\gamma\times W/\gamma$. Afterwards, we flatten the pooled map as a vector of size $1\times HW/\gamma^2$ and feed the vector into 3 consecutive fully-connected (FC) layers. The parameters of these FC layers are denoted as ${\theta}=\left\{\left(\mathcal{W}_{i}, b_{i}\right), i=1, 2, 3\right\}$.
The overall classifier could be denoted as a mapping function: $ \mathcal{G}: X \to \mathcal{G}(X)$ and computed as:
\begin{equation}
    \mathcal{G}(X)=\hat{\sigma}\left(
    \mathcal{W}_{3}{
    \sigma\left(\mathcal{W}_{2}{
    \sigma\left(\mathcal{W}_{1}{\phi(X)}
    +b_{1}\right)}
    +b_{2}\right)}
    +b_{3}\right)
\label{Class_func}
\end{equation}
where $\phi{(\cdot)}$ stands for the combination of the channel averaging, the 2D spatial average pooling and the flattening operation. $\sigma$ is an activation function LeakyReLU and $\hat{\sigma}$ is the Softmax function. $\mathcal{G}(X) \in \{False, True\}$, 
$True$ indicates that the input image is an adversarial sample while $False$ means a clean input. The prediction of classifier determines whether we conduct the subsequent mask operations. If $False$, we skip all the random mask modules. It is feasible because the input and the output of the mask module have the same shape, and are in the same image or feature space. Otherwise, we go through each mask module to mitigate the adversarial perturbations. Note that the input shape in the inference stage might be inconsistent with that in the training stage. To align the inconsistent shapes, we simply insert a spatial adaptive pooling layer after the channel averaging step in the testing stage. 

\subsection{Network Architecture}\label{sec:net}
In this section we first introduce a super-resolution baseline network, and then describe how to incorporate our proposed mask module and classifier into the baseline. Inspired by Newell \textit{et al.}~\cite{newell2016stacked}, we implement an image SR backbone by stacking six hourglass (HG) modules in a sequential manner. For simplicity, Figure~\ref{fig:Network} shows a network architecture with only three HG modules. The hourglass modules have the same architecture but do not share their weights. Two Residual Inception Blocks (RIB) are located in between each two adjacent HG modules. Following Wei \textit{et al.}~\cite{drealsr_wei2020component}, we deploy three component-attentive blocks (CAB) at the end of the 2nd, 4th and 6th hourglass modules respectively. An input image is decomposed of three components: smooth regions, edges and corners. Each CAB only focuses on one of the three components by weighting its SR output with a 2D attentive map. The intermediate output of the three CABs are aggregated to yield the final super-resolved image.

In our proposed method, each hourglass module contains two random frequency mask modules: one at the beginning of the HG module, and the other before the residual blocks.
Our proposed adversarial classifier is placed at the very beginning to predict if the input is adversarial. The classification result is sent to all mask modules and decides if passing through the random mask modules.

\begin{table*}[!t]
	\caption{Comparison between our proposed method and existing super-resolution methods on clean samples and adversarial samples of different intensity. `0/255' denotes clean samples.}
	\label{tab:Tab_sota}
	\small
	\begin{tabular}{|c|c|cc|cc|cc|cc|cc|cc|}
		\hline
		\multirow{2}{*}{\textbf{Method}} & \multirow{2}{*}{\textbf{Scale}} & \multicolumn{2}{c|}{\textbf{0/255}} & \multicolumn{2}{c|}{\textbf{1/255}} & \multicolumn{2}{c|}{\textbf{2/255}} & \multicolumn{2}{c|}{\textbf{4/255}} & \multicolumn{2}{c|}{\textbf{6/255}} & \multicolumn{2}{c|}{\textbf{8/255}} \\ \cline{3-14} 
		&                                 & \textbf{PNSR}  & \textbf{SSIM}  & \textbf{PNSR}    & \textbf{SSIM}    & \textbf{PNSR}    & \textbf{SSIM}    & \textbf{PNSR}    & \textbf{SSIM}    & \textbf{PNSR}    & \textbf{SSIM}    & \textbf{PNSR}    & \textbf{SSIM}    \\ \hline
		\textbf{CARN}                    & \multirow{8}{*}{\textbf{$\times 4$}}    & 28.94          & 0.816          & \textbf{28.72}   & 0.807            & \underline{27.74}      & \underline{0.765}      & \underline{24.86}      & 0.626            & \underline{21.90}      & 0.482            & \underline{20.06}      & \underline{0.389}      \\
		\textbf{SRResNet}                &                                 & 29.12          & 0.823          & 28.17            & 0.796            & 25.38            & 0.690            & 20.23            & 0.428            & 17.25            & 0.259            & 15.69            & 0.172            \\
		\textbf{EDSR}                    &                                 & 29.22          & 0.825          & 28.47            & 0.805            & 26.18            & 0.722            & 21.09            & 0.458            & 18.47            & 0.298            & 16.77            & 0.201            \\
		\textbf{RRDB}                    &                                 & 29.22          & 0.826          & 28.19            & 0.809            & 25.85            & 0.742            & 21.43            & 0.540            & 18.18            & 0.352            & 16.12            & 0.229            \\
		\textbf{RCAN}                    &                                 & \textbf{29.34} & \underline{0.827}    & 28.43            & 0.808            & 25.43            & 0.699            & 20.26            & 0.450            & 17.88            & 0.314            & 16.25            & 0.223            \\
		\textbf{LP-KPN}                  &                                 & 29.13          & 0.823          & 28.16            & 0.790            & 25.20            & 0.663            & 20.40            & 0.417            & 17.68            & 0.274            & 16.12            & 0.198            \\
		\textbf{CDC}                     &                                 & \underline{29.33}    & \textbf{0.828} & 28.37            & \textbf{0.813}   & 26.32            & 0.763            & 23.05            & \underline{0.634}      & 20.70            & \underline{0.501}      & 18.98            & 0.383            \\
		\textbf{Ours}                    &                                 & 29.30          & 0.826          & \underline{28.61}      & \underline{0.809}      & \textbf{27.94}   & \textbf{0.785}   & \textbf{27.65}   & \textbf{0.771}   & \textbf{27.27}   & \textbf{0.762}   & \textbf{26.88}   & \textbf{0.753}   \\ \hline
	\end{tabular}
\end{table*}

\subsection{Adversarial Training}
We employ a stagewise adversarial training strategy to tune our proposed SR network. First, we train the SR backbone network without the mask modules or the classifier. Second, we train the proposed classifier using clean images and adversarial samples that are synthesized with the basic attack~\cite{basicatt_choi2019evaluating} and the SR backbone trained at the first step.
Third, we equip the SR backbone with the trained classifier and all the frequency mask modules to construct our complete SR network. The overall SR network is trained after freezing the classifier and randomly initializing the backbone. Following Wei \textit{et al.}~\cite{drealsr_wei2020component}, We adopt two loss functions, an intermediate one and a gradient-weighted one. To achieve adversarial training, we maximize these loss functions to yield adversarial samples for training the overall proposed SR network.
Note that the adversarial sample classifier is not attacked in the adversarial training.

\section{Experiments}

\subsection{Implementation Details}
We adopt the real-world single image SR dataset RealSR~\cite{realsr_cai2019toward} for evaluation. The dataset contains 595 pairs of LR and HR images. These image pairs are captured by adjusting the focal length of digital cameras, and have been well aligned. Following Cai \textit{et al.}~\cite{realsr_cai2019toward}, 495 pairs are selected for training and 100 pairs are used for testing. Their image sizes are in the range of [700, 3100] and [600, 3500]. $48\times 48$ image patches are cropped for training SR models.
We use the Adam optimizer, exponential decay rate of 0.9, batch size of 16, the initial learning rate of 2e-4. The learning rate is reduced by half after every 1e5 iterations. The maximum number of iterations is 4e5. For the proposed mask module, we set the lower and upper bounds $[r_l, r_u]$ as $[0.43, 0.5]$. For the proposed adversarial classifier, $\gamma$ is set as 3. For adversarial training, we use the basic attack~\cite{basicatt_choi2019evaluating} with 2 iterations and the attack intensity as 6/255. For evaluation, we use the basic attack with 10 iterations, and the intensity $ \alpha \in \{1, 2, 4, 6, 8\}/255$. Peak signal-to-noise ratio (PSNR) and structural similarity (SSIM) are used as the evaluation metrics. Different from Choi \textit{et al.}~\cite{basicatt_choi2019evaluating}, PSNR and SSIM are calculated between the HR ground truths and the super-resolved results of adversarial images.

\subsection{Comparison with the State-of-the-art}
\begin{figure}[!t]
	\centering
	\includegraphics[width=8.cm]{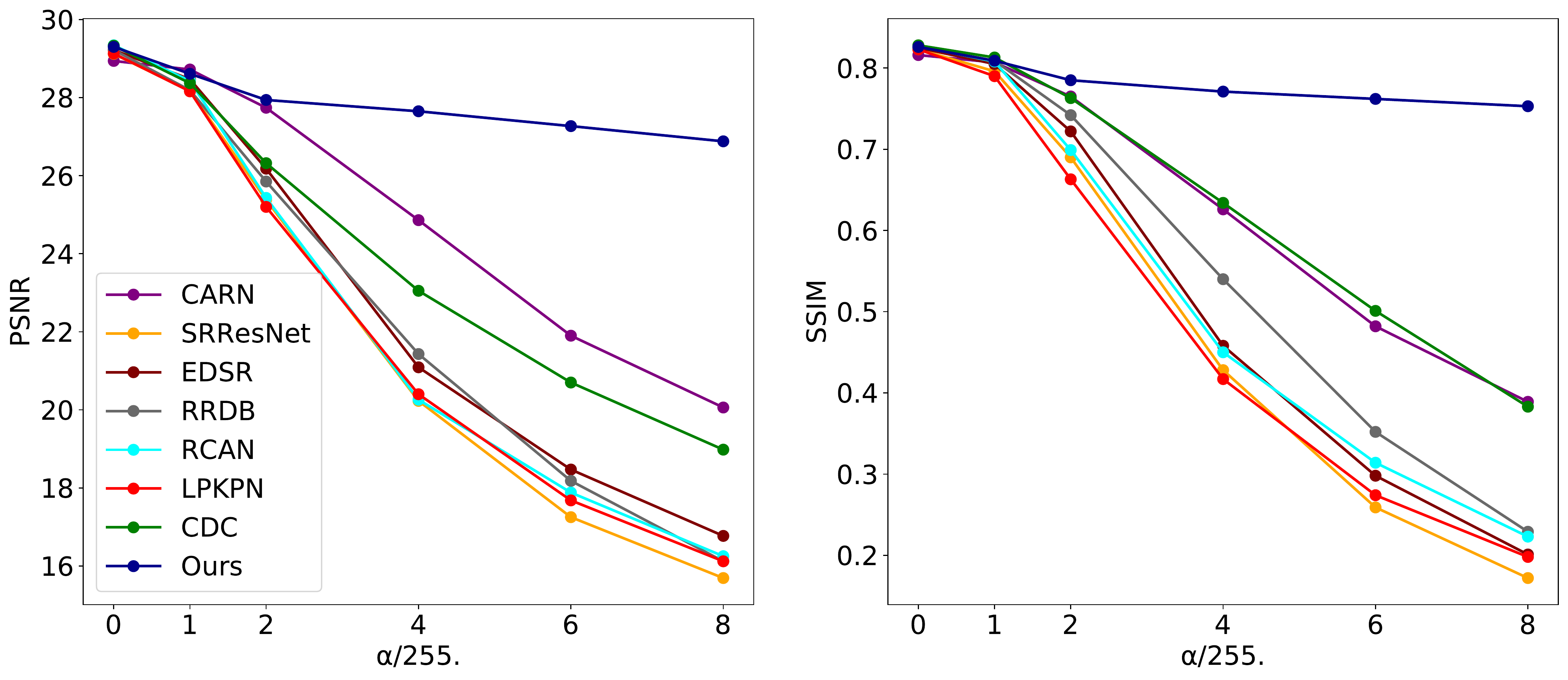}
	\caption{Comparison with the state-of-the-art SR methods against the attacks of different intensity $\alpha$.}
	\label{fig:Sota_pert}
\end{figure} 
We compare our proposed method with existing SR models based on deep CNNs, including
SRResNet~\cite{srres_ledig2017photo}, EDSR~\cite{edsr_lim2017enhanced}, RRDB~\cite{rrdb_wang2018esrgan}, RCAN~\cite{rcan_zhang2018image}, CARN~\cite{carn_ahn2018fast}, LP-KPN~\cite{realsr_cai2019toward}, CDC~\cite{drealsr_wei2020component}.
CARN is a model designed for lightweight image SR, LP-KPN and CDC are developed for real-world image SR. We train these methods on the RealSR dataset and evaluate them from two perspectives below.

\textbf{Qualitative Evaluation} Figure \ref{fig:Sota_image} visually compares the super-solved results of our proposed model with those of existing models. The upper row contains the cropped patches of input images, including a clean LR image on the left and the adversarial LR sample on the right. These adversarial samples are synthesized with the intensity $\alpha = 8/255$ and the iteration number $T = 10$. The middle row is the cropped super-resolved results of clean inputs, while the lower one is the outputs with adversarial samples.
Figure~\ref{fig:Sota_image} shows that our method is able to reconstruct sharp textures and maintain high-level clarity with clean samples. Under the adversarial attack, existing SR models usually produce unreasonable and structured artifacts, which severely affect the quality of super-resolved images. However, such highly repetitive artifacts are not seen in the results produced by our proposed method. This confirms to a certain extent that the proposed mask modules can effectively reduce high-frequency noises.

\textbf{Quantitative Evaluation} Table~\ref{tab:Tab_sota} shows the quantitative results in terms of PSNR \& SSIM. Under the attack intensity of $8/255$, the proposed network achieves the best value of PSNR and SSIM: 26.88 dB and 0.753 respectively. Compared with our baseline CDC~\cite{drealsr_wei2020component}, our proposed network obtains an improvement of 7.9 dB in PSNR and 0.37 in SSIM. For the clean images, the proposed model shows a small performance gap of 0.03 dB in PSNR and 0.002 in SSIM by comparing with the state-of-the-art CDC. Besides, it can be seen in Figure~\ref{fig:Sota_pert} that, as the intensity of adversarial perturbations increases, our proposed model presents smaller decreases than other state-of-the-art methods. Therefore, the above results suggest that our proposed network not only achieves more robust defense than existing real-world SR methods when resisting the white-box attack, but also maintains a competitive performance with clean images.

\subsection{Comparison with Existing Defenses}
\begin{figure}[t]
	\centering
	\includegraphics[width=8.cm]{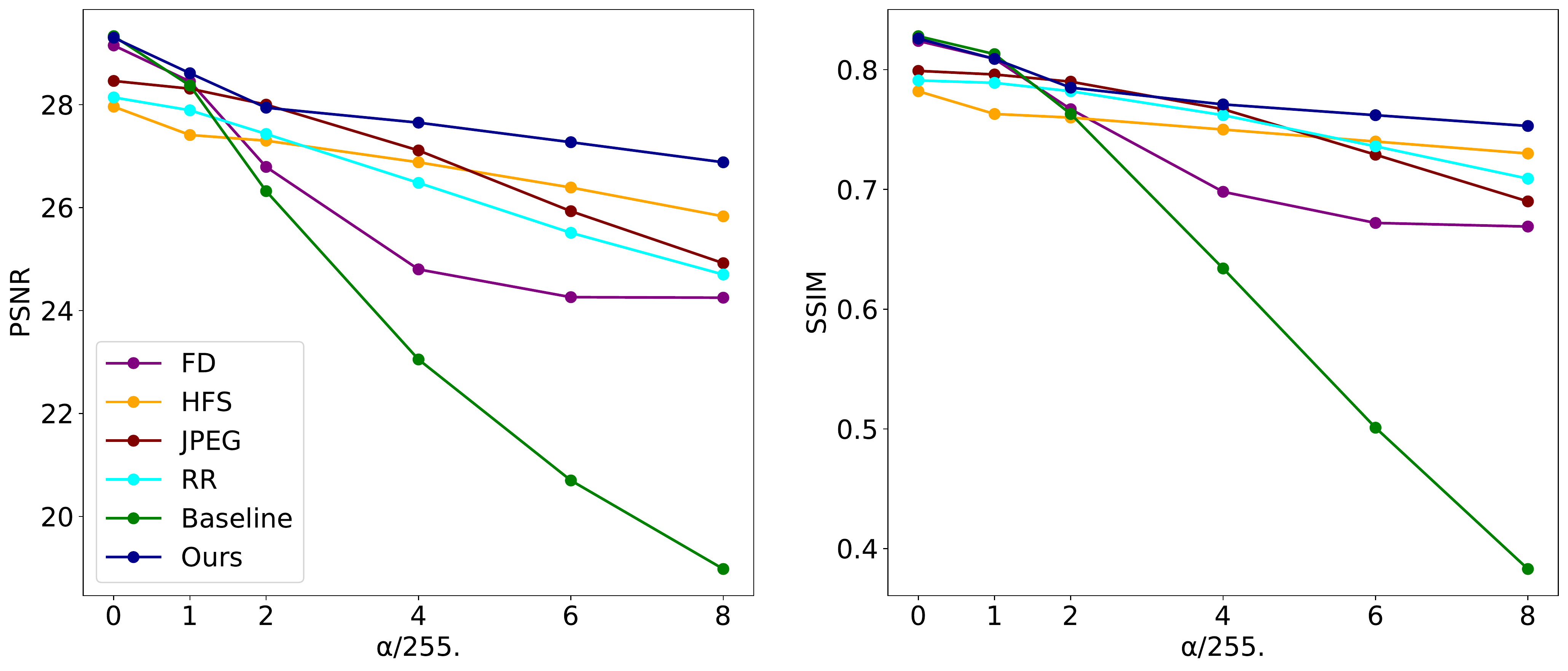}
	\vspace{-1em}
	\caption{Comparison with existing defenses: Feature Denoising (FD), High-Frequency Suppressing (HFS), JPEG, Random Resizing (RR).}
	\label{fig:Defe_pert}
\end{figure}
We further verify the effectiveness of our proposed modules by comparing with other existing defenses, including image compression (JPEG)~\cite{jpeg_dziugaite2016study}, image Random Resizing (RR)~\cite{xie2017mitigating}, a High-Frequency components Suppressing method (HFS)~\cite{zhang2019adversarial} and a Feature Denoising method (FD)~\cite{Xie_2019_CVPR,li2020depthwise}. For the fairness, we adopt CDC~\cite{drealsr_wei2020component} as the baseline that is combined with each of the above defenses respectively in this subsection. We place the JPEG Compression module, the Random Resizing and the High-Frequency Suppressing method in the front of the SR network, following their original setting. We insert four feature denoising blocks at the head of network, the tail of the 2nd, 4th and 6th HG module. Adversarial training strategy is utilized for High-Frequency Suppressing and Feature Denoising, which corresponds to their most effective settings~\cite{zhang2019adversarial,Xie_2019_CVPR}. Figure~\ref{fig:Defe_pert} shows the robustness of all the above defense methods with CDC. It could be observed that our proposed defense obtains the most significant performance at almost every adversarial intensity. With the highest intensity $\alpha = 8/255$, the proposed defense exceeds the second best method by 1.05 dB. The baseline and Feature Denoising could obtains approximate results with our proposed method on clean images. But these two models are far worse than our defense on adversarial samples.

\subsection{Ablation Studies}
\begin{figure}[!t]
	\centering
	\includegraphics[width=8.cm]{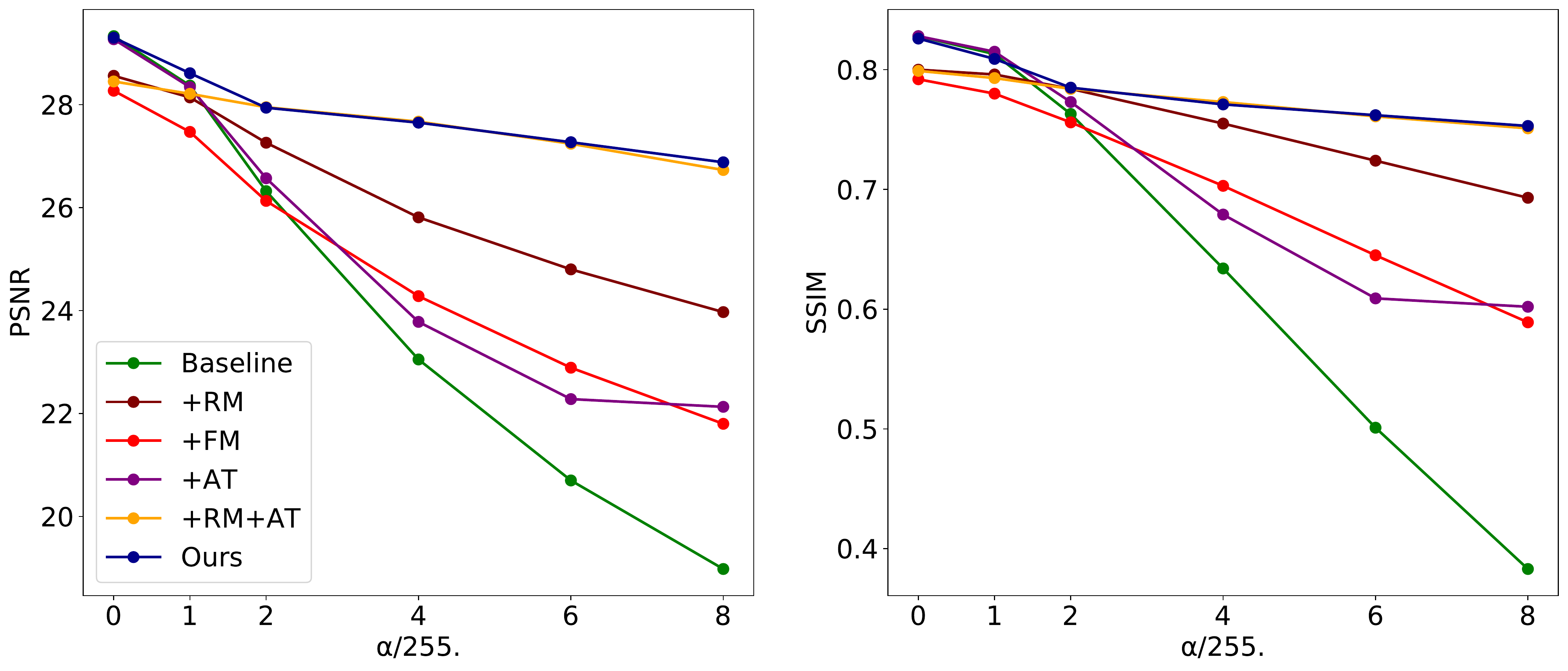}
	\vspace{-1em}
	\caption{Ablation study on the proposed random frequency mask, the proposed adversarial sample classifier and the adversarial training of our method.
	}
	\label{fig:Abla_pert}
\end{figure}

\begin{figure*}[!t]
	\centering
	\includegraphics[width=16cm]{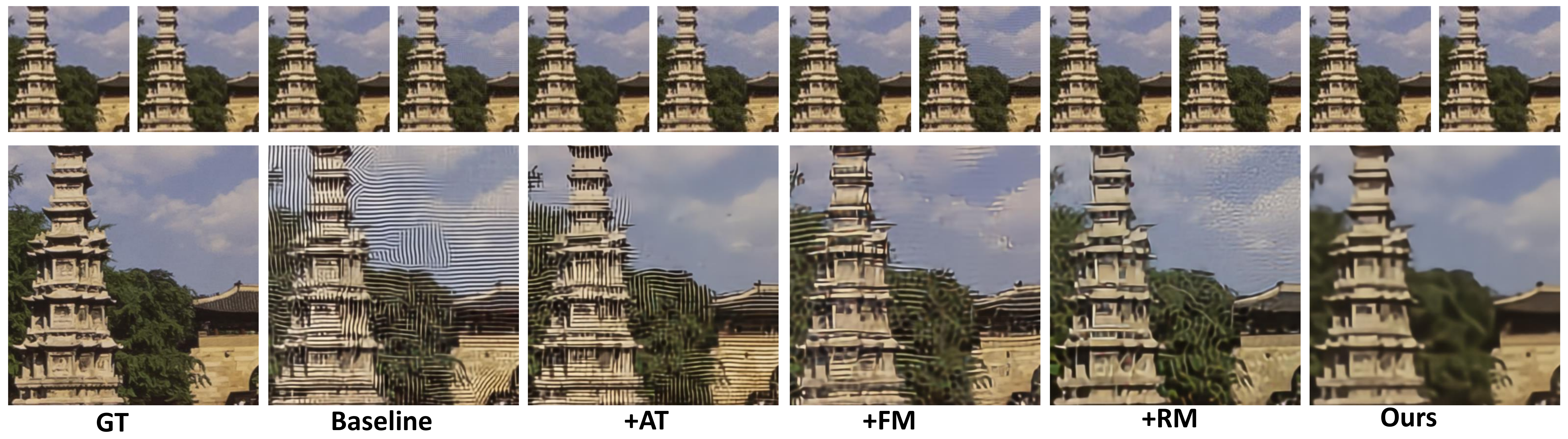}
	\caption{Super-resolved results of the ablation study on adversarial training and our proposed mask module. The first column is the ground truth while the others are for different variants. In each of these columns, the upper left is a clean image while the upper right is the adversarial sample of intensity $\alpha = 8/225$. The bottom row presents the super-resolved results of the adversarial images.}
	\label{fig:Abla_image}
\end{figure*}
\begin{table}[h]
	\caption{Numerical results of the ablation study on our proposed mask module and adversarial classifier in terms of PSNR and SSIM.}
	\label{tab:Tab_abla}
	\setlength{\tabcolsep}{1.1mm}{
		\small
		\begin{tabular}{|c|c|cc|cc|cc|}
			\hline
			\multirow{2}{*}{\textbf{Method}} & \multirow{2}{*}{\textbf{Scale}} & \multicolumn{2}{c|}{\textbf{0/255}} & \multicolumn{2}{c|}{\textbf{4/255}} & \multicolumn{2}{c|}{\textbf{8/255}} \\ \cline{3-8} 
			&                                 & \textbf{PSNR}  & \textbf{SSIM}  & \textbf{PSNR}    & \textbf{SSIM}    & \textbf{PSNR}    & \textbf{SSIM}    \\ \hline
			\textbf{Baseline} & \multirow{6}{*}{\textbf{$\times 4$}}  & \textbf{29.33} & \textbf{0.828} & 23.05 & 0.634 & 18.98 & 0.383  \\
			\textbf{+RM}&     & 28.56 & 0.800 & 25.81 & 0.755 & 23.97 & 0.693  \\
			\textbf{+FM}&     & 28.27 & 0.792 & 24.28 & 0.703 & 21.80 & 0.589  \\
			\textbf{+AT}&     & 29.27 & {\ul 0.828} & 23.78 & 0.679 & 22.13 & 0.602 \\
			\textbf{+RM+AT}&  & 28.45 & 0.799 & \textbf{27.67} & \textbf{0.773} & {\ul 26.73} & {\ul 0.751} \\
			\textbf{Ours} &   & {\ul 29.30} & 0.826 & {\ul 27.65} & {\ul 0.771} & \textbf{26.88} & \textbf{0.753} \\ \hline
		\end{tabular}
	}
\end{table}
We study the effectiveness of the proposed random frequency mask module, the proposed adversarial sample classifier and the adversarial training in our model.
Our proposed network is compared with the baseline (described in Section~\ref{sec:net}), the baseline with the Random frequency Mask (+RM), the baseline with the Fixed frequency Mask (+FM), the baseline with Adversarial Training (+AT), the baseline with Fixed frequency Mask and Adversarial Training (+FM+AT), and the one with Random frequency Mask and Adversarial Training (+RM+AT). The line charts of the above comparison with different attack strengths are plotted in Figure~\ref{fig:Abla_pert}. The +RM setting outperforms the +FM on all values of the attack intensity, which means that the proposed random mask module is more robust than the one with a fixed pre-defined mask. Note that the +RM+AT model is equivalent to removing the proposed classifier from our finally proposed method (denoted as `Ours' in Figure~\ref{fig:Abla_pert}). By comparing the line charts of the +RM+AT model with our method, we find that our proposed model is superior to the +RM+AT model without classifier on clean images or low-intensity adversarial samples. It indicates that the proposed classifier could detect adversarial samples and avoid unnecessary masking to preserve the original image details. 
 
Table \ref{tab:Tab_abla} shows the numerical results of the ablation study. As the table displays, the +RM model surpasses the +FM model by 2.17 dB in PSNR and 0.104 in SSIM against the attack of intensity 8/255. It suggests that the proposed stochastic strategy in our random mask module is effective to improve the super-resolved results, in comparison to the fixed mask module. We verify that adversarial training could enhance the PSNR of the baseline by 3.15 dB with the adversarial samples of intensity 8/255. Besides, our proposed mask module could further boost the baseline with adversarial training by 4.6 dB in PSNR, by comparing the +RM+AT model with the +AT model. Figure~\ref{fig:Abla_image} is the qualitative results of the ablation study, which shows that our proposed method successfully reduces the highly repetitive artifacts in the SR result of the baseline.

\begin{figure}[htb]
	\centering 
	\includegraphics[width=8.cm]{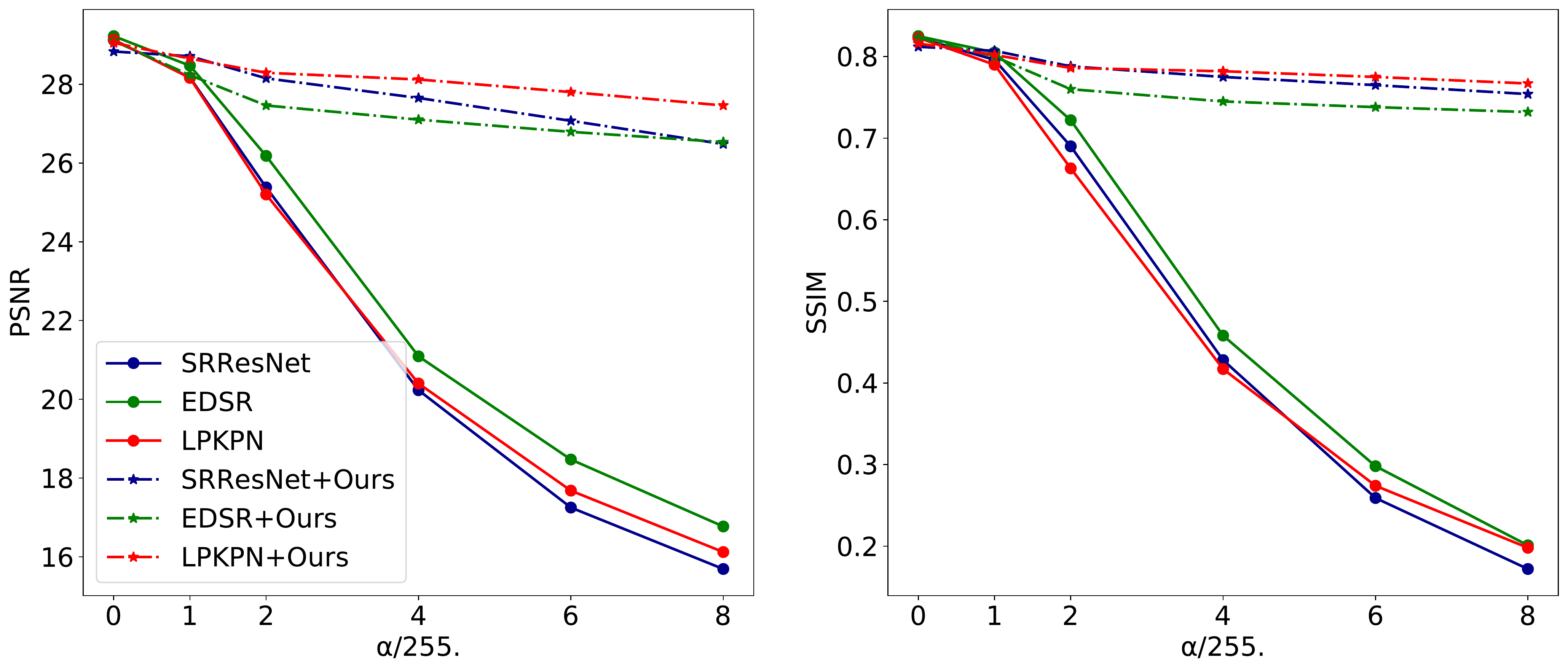}
	\caption{Comparison of robustness with and without our proposed modules in other super-resolution networks. The dotted lines denote results after combining our method.}
	\label{fig:Gener_pert}
\end{figure}
We investigate if our proposed defenses including the random frequency mask and the adversarial classifier are compatible with other super-resolution models. Three state-of-the-art SR networks (including LP-KPN, SRResNet and EDSR) are selected to combine with our proposed modules. For these SR networks, we incorporate a random frequency mask module at their head, and insert a mask module for every four residual blocks. The proposed classifier predicts if the input is adversarial, and determines whether to apply the mask modules.
In Figure \ref{fig:Gener_pert}, the results of three SR networks are shown as solid lines, while the results of adding our proposed modules are dotted lines.
As the figure shows, our proposed defenses significantly enhance the performance against adversarial attacks for the three SR networks. Even with the increase of the attack intensity, the networks with our method still present a stable performance. It indicates that our proposed modules form a general defense framework that has the potential to work with arbitrary SR neural networks.
\vspace{-1em}
\section{Conclusion}
In this paper, we first propose a random frequency mask module that improves the robustness of real-world image super-resolution models. The proposed mask module randomly erases high-frequency components in the discrete cosine transform domain, which is calculated from an image or a convolutional feature map. Considering that the frequency masking operation might be harmful for the original repetitive textures in an input image, we further develop an adversarial sample classifier that avoids unnecessary masking the high-frequency details by detecting adversarial inputs. The proposed mask modules are only activated when the input image is very likely to be attacked. We experimentally show that our proposed method not only defends a wide range of existing SR networks against white-box attacks, but also maintains competitive performance with clean images. Overall, we introduce a general defense framework for real-world image super-resolution and the proposed framework may be extended to other robust applications in future.
\begin{acks}
This work is supported by Key-Area Research and Development Program of Guangdong Province [2020B0101350001], National Natural Science Foundation of China under Grant No.61976250, No.U1811463 and No.62006253, and Guangzhou Science and technology project under Grant No.202102020633.
\end{acks}
\bibliographystyle{ACM-Reference-Format}
\bibliography{main}

\end{document}